\pdfoutput=1
\documentclass[11pt]{article}

\usepackage[]{acl}
\usepackage{times}
\usepackage{latexsym}

\usepackage[T1]{fontenc}
\usepackage[utf8]{inputenc}

\usepackage{microtype}

\usepackage{inconsolata}
\usepackage{multirow}
\usepackage{tabularray}
\UseTblrLibrary{booktabs}
\usepackage{siunitx}
\usepackage{xcolor}
\usepackage{tabularx}
\usepackage{array}
\usepackage{lscape}
\usepackage{arydshln}
\usepackage{graphicx}
\usepackage{fontawesome}

\newcolumntype{P}[1]{>{\centering\arraybackslash}p{#1}}
\newcolumntype{Y}{>{\centering\arraybackslash}X}
\def\0{\phantom{0}} 
\def\+{\phantom{+}}

\title{Can Large Language Models Understand Context?}

\author{
    Yilun Zhu$^{1}$\thanks{\rule{1ex}{0ex}Work performed during an internship at Apple.}, Joel Ruben Antony Moniz$^{2}$, Shruti Bhargava$^{2}$, Jiarui Lu$^{2}$\\
    \textbf{Dhivya Piraviperumal$^{2}$}, \textbf{Site Li$^{2}$}, \textbf{Yuan Zhang$^{2}$}, \textbf{Hong Yu$^{2}$}, \textbf{Bo-Hsiang Tseng$^{2}$} \\
    $^1$Department of Linguistics, Georgetown University \\
    $^2$Apple \\
    \texttt{yz565@georgetown.edu} \\
    \texttt{\{joelrubenantony\_moniz, shruti\_bhargava, jiarui\_lu, dhivyaprp\}@apple.com} \\
    \texttt{\{site\_li, yzhang73, hong\_yu, bohsiang\_tseng\}@apple.com}
}

\begin{document}
\maketitle

\begin{abstract}
Understanding context is key to understanding human language, an ability which Large Language Models (LLMs) have been increasingly seen to demonstrate to an impressive extent. However, though the evaluation of LLMs encompasses various domains within the realm of Natural Language Processing, limited attention has been paid to probing their linguistic capability of understanding contextual features. This paper introduces a context understanding benchmark by adapting existing datasets to suit the evaluation of generative models. This benchmark comprises of four distinct tasks and nine datasets, all featuring prompts designed to assess the models' ability to understand context. First, we evaluate the performance of LLMs under the in-context learning pretraining scenario. Experimental results indicate that pre-trained dense models struggle with understanding more nuanced contextual features when compared to state-of-the-art fine-tuned models. Second, as LLM compression holds growing significance in both research and real-world applications, we assess the context understanding of quantized models under in-context-learning settings. We find that 3-bit post-training quantization leads to varying degrees of performance reduction on our benchmark. We conduct an extensive analysis of these scenarios to substantiate our experimental results.\footnote{
The code is publicly available at \url{https://github.com/apple/ml-llm-contextualization-eval}.}
\end{abstract}

\section{Introduction}

Discourse understanding, as one of the fundamental problems in NLP, focuses on modeling linguistic features and structures that go beyond individual sentences \cite{joty-etal-2019-discourse}. Understanding discourse requires resolving the relations between words/phrases (coreference resolution) and discourse units (discourse parsing and discourse relation classification) in the previous context, identifying carry-over information for the following context (dialogue state tracking), and recognizing discourse-specific phenomena (ellipsis).

LLMs have garnered substantial attention from both academia and the industry due to their remarkable capability in comprehending language and world knowledge. Their unparalleled performance across a diverse range of benchmarks and datasets has firmly established their significance in a relatively short period of time. As LLMs continue to push the boundaries of scale and capability, the evaluation of their multifaceted abilities becomes an equally vital endeavor. Consequently, the development of robust evaluation methodologies to assess specific aspects of LLMs becomes imperative. In addition, these methodologies should focus on helping achieve a comprehensive understanding of their advancement while clearly delineating their limitations.
However, recently published LLMs, such as OPT \cite{zhang2022opt}, LLaMA \cite{touvron2023llama} and GPT-4 \cite{openai2023gpt4}, are only evaluated on limited benchmarks, and have a significant drawback: they neglect the inclusion of discourse-related datasets for evaluation, thereby limiting the comprehensive assessment of their language understanding capabilities.
% \andy{not clear to me the major diff between the two}

% \begin{table}[t!b]
%     \centering\small
%     \aboverulesep=0ex
%     \belowrulesep=0ex
    
%     \begin{tabularx}{0.48\textwidth}{c|c|c|>{\centering\arraybackslash}X}
%     \toprule
%     \rule{0pt}{2ex}Type & Task & Dataset & Context \\ 
%     \midrule
%     \rule{0pt}{2.2ex}\multirow{5}{*}{Doc} & \multirow{2}{*}{Coreference} & WSC273 & Nominal \&    \\ \cline{3-3}
%     \rule{0pt}{2.2ex} && OntoNotes & eventual reference \\ \cline{2-4} 
%     \rule{0pt}{2.2ex} & \multirow{2}{*}{Discourse} & \multirow{2}{*}{PDTB-3} & Relations between discourse units \\ 
%     \midrule
%     \rule{0pt}{2.2ex}\multirow{6}{*}{Dial.} & DST & \multirow{2}{*}{MultiWoz}  & Entity carryover within context \\ 
%     % \cline{3-3}
%     % \rule{0pt}{2ex} && SGD &  \\ 
%     \cline{2-4} 
%     \rule{0pt}{2.2ex} & \multirow{5}{*}{\begin{tabular}[c]{@{}c@{}}Query\\ Rewrite\end{tabular}} & MuDoCo & \multirow{5}{*}{Ellipsis and reference}          \\ \cline{3-3}
%     \rule{0pt}{2.2ex} && QReCC &  \\ \cline{3-3}
%     \rule{0pt}{2.2ex} && InCar &  \\ \cline{3-3}
%     \rule{0pt}{2.2ex} && GECOR &  \\ \cline{3-3}
%     \rule{0pt}{2.2ex} && CANARD &  \\
%     \bottomrule
%     % \hline
%     \end{tabularx}
    
%     \caption{Tasks and datasets in the context understanding benchmark.}
%     \label{tab:task_table}
% \end{table}

\begin{figure*}[t!]
    \centering
    \includegraphics[width=\textwidth]{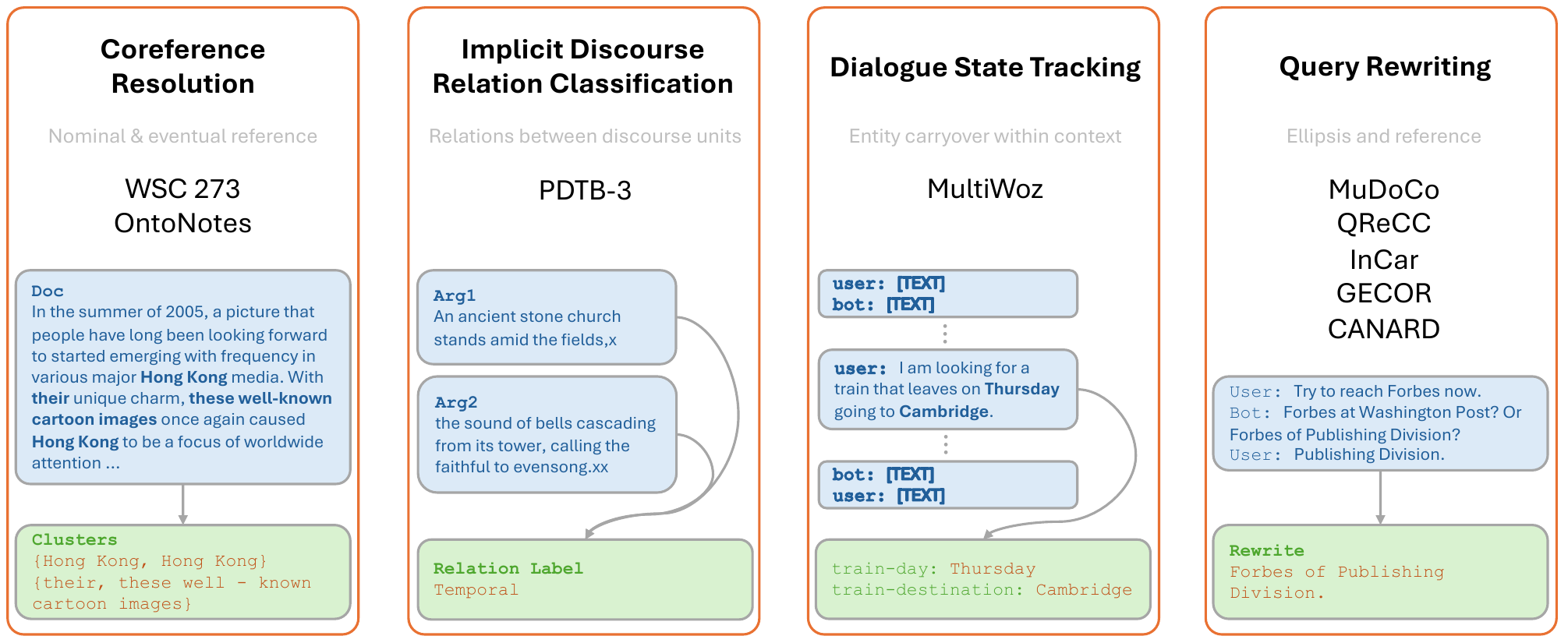}
    \caption{Tasks and datasets in the context understanding benchmark.}
    \label{fig:benchmark}
\end{figure*}

To provide a comprehensive evaluation, plenty of benchmarks and datasets address various facets of language understanding, including benchmarks that delve into common sense knowledge \cite{hendryckstest2021, kwiatkowski-etal-2019-natural}, as well as linguistic capabilities like sentiment analysis, natural language inference, summarization, text classification, and more \cite{bang2023multitask, liang2022holistic}. 
% Previous research has also placed specific emphasis on assessing the efficacy of LLMs within distinct contextual comprehension datasets, such as coreference resolution \cite{le2023large} and dialogue-state tracking (DST) \cite{heck-etal-2023-chatgpt}. 
These general benchmarks and specific dataset evaluations exhibit certain limitations. Despite the requirement for contextual information in these benchmarks to effectively tackle tasks (for example, sentiment analysis requires an understanding of polarities within the given text), none of these benchmarks cater to tasks that demand a nuanced comprehension of linguistic features within a provided context.
% Secondly, the attention of specific dataset evaluations for contextual comprehension is paid to large and powerful models, such as ChatGPT \cite{openai2022chatgpt}, inadvertently neglecting the potential of smaller models to achieve competent performance on these tasks 
% \andy{note sure if we want to point out the deficiencies of these datasets as we are still using them afterwards}

On the other hand, recent LLMs, by virtue of possessing billions of parameters, have led to an exponential surge in computational and storage costs \cite{brown2020language}, which hinders the deployment of large models to personal devices and restricts the on-device performance of language understanding tasks. To address this challenge, model compression methods, which can reduce memory and disk requirements of both model training and inference, have gained attention. 
Existing compression techniques, such as 3-bit quantization \cite{frantar-gptq}, have demonstrated the potential to reduce model sizes
% to nearly 30\% of the original dense model, 
with only marginal performance trade-offs.
% \andy{is 30\% estimated by paper?}
However, the evaluation of quantization methods suffers from two deficiencies. Firstly, quantization methods are primarily evaluated on limited benchmarks and datasets, such as Lambada \cite{paperno-etal-2016-lambada}, ARC \cite{boratko-etal-2018-systematic}, PIQA \cite{tata2003piqa}, BoolQ \cite{clark-etal-2019-boolq}, and StoryCloze \cite{mostafazadeh-etal-2017-lsdsem}. It is not yet clear whether large, compressed models out- or under-perform their smaller counterparts when understanding context. Secondly, previous work has not delved into a linguistic analysis to identify where the model efficacy wanes.
% \andy{same here, not clear to me the major diff between the two}

Given the above shortcomings, this paper evaluates LLMs on a context understanding benchmark constructed from varied discourse understanding datasets.
% , many of which were originally designed for classification tasks.
% \andy{qr are not cls. task}
% \yz{Done, remove the second part}
We conduct an extensive analysis of LLM performance on this benchmark, including models of varying sizes and those subjected to compression techniques, aiming to provide a more comprehensive understanding of context understanding capability of the LLMs.
The contributions of this paper can be summarized as follows:

\begin{itemize}
\item Our work introduces a contextual understanding benchmark, including four tasks, for the evaluation of LLMs. We also present prompts designed for in-context learning on each task.
% \andy{highlight the design of the prompts proposed?}
% \yz{Done}

\item We evaluate LLMs of varying sizes from different model families and provide an analysis on these models' capability for context understanding.
% \andy{from various model families}
% \yz{Done}
% demonstrating the weak capability of LLMs in understanding context.
% \andy{providing a unique insight on models' capability on these discourse understanding tasks?}
% \yz{Done}
% \andy{swap 2nd the 3rd points}

\item We evaluate post-training compressed models in ICL settings and conduct an analysis of the reduction in context understanding capability compared to dense models.
% showing that the contextual comprehension capability significantly drops under 3-bit quantization.
% \andy{not really right? many of the tasks do not have drops}
% \yz{Done, with more abstract expression}
\end{itemize}

\section{Related Work}
\subsection{In-context Learning Evaluation}
The paradigm of ICL  \cite{NEURIPS2020_1457c0d6} is rapidly gaining importance. Studies have demonstrated that the generalization of LLMs to various downstream NLP tasks, such as 
% \jm{I assume this stands for Massive Multitask Language Understanding?}
 MMLU \cite{hendrycks2021measuring}, is significantly enhanced when provided with a small number of examples as prompts \cite{NEURIPS2020_1457c0d6, chowdhery2022palm, hoffmann2022training, rae2022scaling, anil2023palm, touvron2023llama, openai2022chatgpt, openai2023gpt4}. 
Recent research has extensively evaluated the performance of LLMs across a spectrum of language-related tasks, spanning from text generation to understanding input sequences. This assessment contains a wide array of benchmarks, including SUPER-GLUE \cite{NEURIPS2019_4496bf24, laskar-etal-2023-systematic}, and tasks such as question answering, information retrieval, sentiment analysis \cite{bang2023multitask, liang2022holistic}, dialogue \cite{heck-etal-2023-chatgpt}, and text classification \cite{yang2023large}.

\subsection{Model Compression for LLMs}
Model compression techniques can be broadly categorized into three main approaches: compression during training, compression associated with fine-tuning, and post-training methods.
In terms of quantization during training, this technique enables LLMs to adapt to low-precision representations during the training process \cite{liu2023llmqat}. 
Model compression with fine-tuning involves quantization awareness into the fine-tuning stage \cite{kim2023memoryefficient, dettmers2023qlora}.
Post-training techniques, on the other hand, are applied after the completion of an LLMs training phase and typically involve the use of calibration data. This category comprises two primary approaches: pruning, which removes redundant or non-salient weights to induce sparsity \cite{frantar-sparsegpt}, and quantization, which employs low-precision numeric representations of weights and activations \cite{nagel-etal-2020-adaptive, frantar-gptq, yuan2023rptq}.
Prior research shows that quantization outperforms pruning in several settings \cite{kuzmin2023pruning}, thus in this work, we focus on model quantization and its impact on the selected context-aware tasks.
% \andy{simply put we focus on ICL evaluation, so techniques involving weights updates are not suitable}
% \yz{Done, move to Sec 4.3}
% Since this paper targets at contextual comprehension evaluation under ICL settings, quantization-aware training and fine-tuning techniques are not considered.
% Since quantization-aware training consumes a significant amount of GPU resources, and because fine-tuning techniques influence the evaluation of LLMs' context understanding by incorporating benchmark-related data, we opt for the post-training method 
% \jm{Should we specify also why we choose quantization over pruning?  }
% \yz{limited by paper length?}
% \shruti{can cite works like https://arxiv.org/pdf/2307.02973.pdf}
% \yz{Done}
% to evaluate compressed LLMs on our contextual comprehension benchmark.

\section{Task Selection \& Design}
% \yz{Trimmed: coreference resolution -> CR; query rewriting -> QR}

Our contextual understanding benchmark includes four tasks with nine 
% \andy{-> 9}
% \yz{Done}
datasets, as presented in Figure \ref{fig:benchmark}.
% \andy{as shown in Table 1}
% \yz{Done}
In the following sections, we provide detailed explanations of each task and the corresponding datasets, along with the designed prompts for ICL evaluations.
% Considering some tasks are not suited for generative models, we carefully design these tasks as well as instructions and prompts to guarantee that the task is not the major factor hindering the performance of LLMs.
% \andy{what you mean by task is not the major factor?}
% \andy{In the following sections we explain each task and the corresponding datasets with designed prompts for ICL evaluation.}
% \yz{Done}
% Detailed examples for each task design can be found in Appendix \ref{appendix:task_design}.
% \andy{move and adapt this sentence to the first time you show prompt (i.e., when you use table 2)}
% \yz{Done}

\subsection{Coreference Resolution} \label{subsec:coref}
% \andy{some can be cut when trimming}
%%%%%% Trimmed by Yilun %%%%%
% Coreference resolution (CR), as a key component of Natural Language Understanding (NLU), is the task of forming connections and understanding relations among various segments of a text. This enables the interpretation of pronouns, nouns, and other expressions in relation to the entities they are referring to. This task \yz{Trimmed suggestion: remove the above and starts with `The coreference resolution task (CR)'}

The coreference resolution (CR) task contributes to achieving a coherent understanding of the overall meaning conveyed within the text. Thus, it plays a critical role in diving into language models' capability to grasp coreference relations as well as contextual nuances within documents. We select two coreference datasets: WSC273 \cite{wsc} and OntoNotes 5.0 \cite{pradhan-etal-2013-towards}. 

\begin{table}[t!b]
    \centering\small
    \aboverulesep=0ex
    \belowrulesep=0ex
    \begin{tabularx}{0.48\textwidth}{X}
        \toprule
        % \hline
        \rule{0pt}{2.2ex}\textbf{Instruction}: Please carefully read the following passages. For each passage and the options, you must identify which option the mention marked in *bold* refers to. If the marked mention does not have any antecedent, please select ``no antecedent''. \\
        \textbf{Context}: ... To express *its* determination ... the Chinese securities regulatory department ... this stock reform ... \\
        \textbf{Choices}:\\
        A. no antecedent \\
        B. the Chinese securities regulatory department \\
        C. this stock reform \\
        ... \\
        \textbf{Question}: What does *its* refer to? \\
        \textbf{Answer}: \textit{B} \\
        \bottomrule
        % \hline
    \end{tabularx}
    \caption{An OntoNotes example of prompt and \textit{answer}.}
    \label{tab:coref_design}
\end{table}
 
% \andy{we didn't explain WSC 273?}
% \yz{Done}
WSC273, which contains the first 273 examples from the Winograd Schema Challenge, is a dataset that requires the system to read a sentence with an ambiguous pronoun and select the referent of that pronoun from two choices.
OntoNotes is a human-annotated corpus of documents annotated with multiple layers of linguistic information including syntax, propositions, named entities, word sense, and in-document 
coreference. As it is one of the most frequently used datasets for training coreference models, prior research has achieved significant advancements under the supervised fine-tuning paradigm \cite{lee-etal-2017-end, joshi-etal-2020-spanbert, bohnet-etal-2023-coreference}. However, these model designs cannot be extended to generative models under ICL settings.
% design of the classification task proves unsuitable for extension to generative models and few-shot learning contexts
% \jm{this wasn't very clear to me, why? Maybe we need to elaborate/clarify here?}
% \yz{Refine the sent}
Recently, \citet{le2023large} have leveraged document templates for LLMs; however, their evaluation is confined to prominent models such as InstructGPT \cite{ouyang2022training}, neglecting the fact that smaller models lack the generative capacity required to accomplish such tasks. Due to these limitations, we propose a novel multiple-choice task design. In this design, we provide the mentions and evaluate the model on resolution. Each option represents a potentially markable span.\footnote{Considering the inferior performance of small models on the mention detection task, we utilize gold markable spans coreference linking.
% \jm{maybe we could say "gold mention coreference linking spans"?}\yz{change to ``gold markable spans''}
} Table \ref{tab:coref_design} presents an example of the input to the model\footnote{Detailed examples for each task design can be found in Appendix \ref{appendix:task_design}.}.
% \andy{all tables use tbh}
% \yz{Done}
% \andy{include, for example 1 line in each part (Given a paragraph and ...), "[Instruction]" looks like it's what it is in prompt; refer to Appendix for complete prompt}
% \yz{Done}
The entire prompt consists of five parts: (1) an instruction that provides guidance to the model for the task, (2) a document containing plain text with a selected mention span highlighted using a bold symbol, (3) a list of choices, which includes all the gold mentions present in the document, (4) a question that directs the model's attention, and (5) a guiding word \textit{answer} that prompts for the output. We experiment with multiple instructions and prompts and provide the one with the best performance.
Linking scores are computed for each question and the results are subsequently aggregated for evaluation.
% \andy{elaborate a bit more on how to select}
% \yz{Done}
We utilize the official evaluation metrics from the CoNLL-2012 shared task \cite{pradhan-etal-2012-conll}, which employs the CoNLL F1 score, derived from the averaging of three coreference metrics: MUC, B$^3$, and CEAF$_{\phi4}$.

\subsection{Dialogue State Tracking}

\begin{table}[t!b]
    \centering\small
    \aboverulesep=0ex
    \belowrulesep=0ex
    \begin{tabularx}{0.48\textwidth}{X}
        \toprule
        % \hline
        \rule{0pt}{2.2ex}\textbf{Ontology}: \\
        \{``slots'': \{``restaurant-pricerange'': ``price budget for the restaurant'', ... \}, \\
        ``categorical'': \{``restaurant-pricerange'': [`cheap', `expensive', `moderate'], ...\} \}\\
        
        \textbf{Instruction}: Now consider the following dialogue between two parties called the ``system'' and ``user''. Can you tell me which of the ``slot'' was updated by the ``user'' in its latest response to the ``system''? Present the updates in JSON format. If no ``slots'' were updates, return an empty JSON list. If you encounter ``slot'' that was requested by the ``user'' then fill them with ``?''. If a user does not seem to care about a discussed ``slot'' fill it with ``dontcare''.\\
        \textbf{[Previous Dialogue State]}\\
        \textbf{[Conversation]}: \\
        ``system'': ``''\\
        ``user'': ``I'm looking for a moderately priced place to eat that's in the centre of town.''\\
        \textbf{Output}: \textit{\{``restaurant-pricerange'': ``moderate'', ``restaurant-area'': ``centre''\}}\\
        \bottomrule
        % \hline
    \end{tabularx}
    \caption{A DST example of prompt and \textit{answer}. 
% \andy{Put output as well, separate from prompt, for all tasks.}
% \yz{Done}
}
    \label{tab:dst_design}
\end{table}

% \andy{andy todo: refine a bit}
Dialogue state tracking (DST) is an important task in the area of task-oriented dialogue (TOD) modeling \cite{young2013pomdp}, where the dialogue agent tracks the key information provided by the user as the conversation progresses. Table ~\ref{tab:dst_design} provides an example from MultiWOZ \cite{budzianowski-etal-2018-multiwoz} where the user expresses the constraints when looking for a restaurant. The output of DST is typically maintained in slot-value pair format.
% involves continuously monitoring and updating the internal representation of a conversation's current context, including user intents, preferences, and system actions. It is an important component of the proposed contextual comprehension benchmark because the task enables the system to maintain context across turns (even between turns with long dependencies), accurately interpret user requests, and refer to running example here without real output.
% \andy{Refer to running example here without real output}

Previous research has explored ICL capabilities on MultiWOZ and demonstrated promising results compared to fine-tuning models \cite{hu-etal-2022-context, heck-etal-2023-chatgpt}. However, these studies either involve partial training or are untested with smaller and quantized models.
% Therefore, we have adopted the straightforward and simplified ICL approach proposed by \citet{heck-etal-2023-chatgpt} for both the MultiWOZ V2.2 \cite{zang2020multiwoz} and the Schema-Guided Dialogue (SGD) \cite{rastogi2020towards} datasets.
Here we adopt a straightforward and simplified ICL approach proposed by \citet{heck-etal-2023-chatgpt}, and test it on MultiWOZ v2.2 \cite{zang2020multiwoz}.
% This approach uses a template that excludes the influence of other factors and is consistent across models.
% \noindent
The prompt to the model consists of domain knowledge from ontology, an instruction, previous dialogue state (the belief state accumulated until the previous user turn) and the conversation proceeding to the current turn. The ontology could be lengthy if considering all domains in the dataset. Thus, given the input length constraint of LLMs, only the knowledge relevant to the conversation is provided.
% Considering the model input length, we only feed domain knowledge relevant to the current dialogue in a structured format, the previous dialogue state, and the conversation proceeding to the current user turn into the prompt.
% As shown in Table \ref{tab:dst_design}, the conversation is within the restaurant domain, and therefore, we exclusively provide the model with knowledge related to restaurants from entire ontology. The previous state represents the prediction from the preceding turn pair. 
Following literature, we report joint goal accuracy (JGA) \cite{mrkvsic2017neural} for evaluating the performance of DST.

\subsection{Implicit Discourse Relation Classification}
Discourse demonstrates its importance beyond individual sentences, which emphasizes the ways in which different segments of a text interconnect and structure themselves to convey a coherent and meaningful message. The PDTB-3 corpus, as introduced by \citet{webber2019pdtb3}, annotates implicit discourse relations across elementary discourse units (EDUs)\footnote{EDU refers to the smallest segment of a text that conveys a complete and coherent meaning within larger discourse.}. These relations imply connections between EDUs and may be made explicit by inserting a connective.
%%%%% Trimmed by Yilun %%%%%
% This corpus encompasses two types of discourse relations: explicit relations, where a connective such as \textit{and} or \textit{but} appears between two EDUs, and implicit relations, where a connection could be inferred between them and potentially made explicit by inserting a connective.
Within the context of the understanding benchmark, we opt for the implicit discourse relation classification task for two primary reasons. Firstly, the order of the two EDUs is provided, enabling the model to directly utilize this information. Secondly, the connective triggering the relation is implicit, increasing the task's complexity.
In this task, two EDUs are fed as input, and the objective is to correctly identify the relation between them. Due to the nuanced differences between each relation and the demand for annotators with rich linguistic knowledge and extensive annotation training, the classification task poses challenges to fine-tuned classification models. 

\begin{table}[t!b]
    \centering\small
    \aboverulesep=0ex
    \belowrulesep=0ex
    \begin{tabularx}{0.48\textwidth}{X}
        \toprule
        \hline
        \rule{0pt}{2.2ex}\textbf{Instruction}: Given two arguments and a list of connective words, please select the most likely connective between two arguments. \\
        \textbf{[Relation Description]}\\
        \textbf{Input}: \\
        Arg 1: Amcore, also a bank holding company, has assets of \$1.06 billion. \\
        Arg 2: Central's assets are \$240 million. \\
        \textbf{Question}: What is the connective that best describes the relation between two arguments? \\
        \textbf{Choices}:\\
        A. Temporal  B. Contingency C. Comparison D. Expansion \\
        \textbf{Answer}: \textit{C}\\
        \bottomrule
        % \hline
    \end{tabularx}
    \caption{A PDTB example of prompt and \textit{answer}.}
    \label{tab:pdtb_design}
\end{table}

The PDTB3 corpus classifies discourse relations into four categories - \texttt{Temporal}, \texttt{Contingency}, \texttt{Comparison}, and \texttt{Expansion}. We convert this task into a multiple-choice question and experiment with \emph{classes} as options. In the \emph{classes} scenario, the task offers four options, each representing a distinct discourse relation class.
% with two types options: \emph{classes} and \emph{connectives}. In the \emph{classes} scenario, the task offers four options, with each representing a distinct discourse relation class. In the \emph{connectives} approach, we follow previous work that fine-tunes the model \cite{zhou-etal-2022-prompt-based, xiang-etal-2023-teprompt}, requiring the model to predict a connective from a predetermined set and to subsequently map this prediction to the corresponding relation.
% \andy{unclear to me}
% \yz{Done, remove connectives}
% Experimental results show that the class scenario with an explanation of each class provided in the prompt achieves the best scores.
% \andy{unclear to me}
% \yz{Done, same as above}
Table \ref{tab:pdtb_design} exhibits the components of the prompt.
% \andy{again, don't use "[]" but use 1 sentence}
% \yz{Done}
It includes an instruction at the beginning, followed by a concise description of each relation, a context with two arguments, a question along with answer choices, and a trigger word.
We evaluate each model's performance on this dataset using accuracy as the metric.

\subsection{Query Rewriting}

While document-based CR (OntoNotes, Section \ref{subsec:coref})
% \andy{put link to Sec3.1}
% \yz{Done}
covers various types of coreference relations across multiple genres, it does not allow the ability to evaluate certain aspects which are important to understand context. Firstly, the
% \jm{I think we should say "the coreference resolution task" rather than "our benchmark", because this is more a short-coming of OntoNotes?}
CR task typically focuses on document-based coreference chains, neglecting mention resolution in dialogues. Secondly, ellipsis, which is the omission of one or more words from a clause, is a crucial linguistic phenomenon in speech and conversation. It is essential for language models to grasp and accurately identify ellipses within context. Incorporating these features into the benchmark is thus pivotal when evaluating context understanding.

\begin{table}[t!b]
    \centering\small
    \aboverulesep=0ex
    \belowrulesep=0ex
    \begin{tabularx}{0.48\textwidth}{X}
        \toprule
        % \hline
        \rule{0pt}{2.2ex}\textbf{Instruction}: Rewrite the last query following interaction into a well-formed, context independent query. Resolve any disfluencies or grammatical errors in the query. \\
        \textbf{Input}: \\
        User: Try to reach Forbes now . \\
        Bot: Forbes at Washington Post ? Or Forbes of Publishing Division ? \\
        User: Publishing Division . \\
        \textbf{Rewrite}: \textit{Forbes of Publishing Division}\\
        \bottomrule
        % \hline
    \end{tabularx}
    \caption{A query rewriting example of prompt and \textit{answer}.}
    \label{tab:qr_design}
\end{table}

Query Rewriting (QR) is a task of rewriting the last utterance of a user in a conversation into a context-free, independent utterance that can be interpreted without dialog context. It requires the model to identify the entity or events references from
% \jm{maybe we should replace "mentioned in" with "referenecs from"?}
context and further generate a complete utterance with resolved coreference or ellipsis.
% \jm{"that references the previous context directly"}

We incorporate five QR datasets in the proposed benchmark: MuDoCo with QR annotations \cite{martin-etal-2020-mudoco, tseng2021cread}, QReCC \cite{anantha-etal-2021-open}, InCar \cite{regan2019incar}, GECOR \cite{quan-etal-2019-gecor}, and CANARD \cite{elgohary-etal-2019-unpack}. These datasets span multiple genres and domains in dialogues. We experiment with various prompts used for fine-tuning models and present the results with the best selections.
% \jm{"the results with the best selections" -> "the best-performing results"?}
Table \ref{tab:qr_design} presents a concise prompt comprising an instruction along with context for each dialogue.
To assess the quality of generated queries, we follow the metrics from previous research, particularly BLEU \cite{papineni-etal-2002-bleu} and ROUGE \cite{lin-2004-rouge}.

\begin{table*}[]
    \centering\small
    \aboverulesep=0ex
    \belowrulesep=0ex
    \begin{tabular}{c|c|c|cccc|ccc|c|c}
    \toprule
    % \hline\hline
    \rule{0pt}{2.2ex}\multirow{2}{*}{Task} & \multirow{2}{*}{Dataset} & \multirow{2}{*}{Metrics} & \multicolumn{4}{c|}{OPT} & \multicolumn{3}{c|}{LLaMA} & GPT & \multirow{2}{*}{FT} \\ \cline{4-11}
    \rule{0pt}{2.2ex}& & & 125M & 350M & 1.3B & 2.7B & 7B   & 13B   & 30B   & 3.5-turbo \\ 
    \midrule
    % \hline
    \rule{0pt}{2.2ex}\multirow{5}{*}{CR} & WSC273 & Acc  & 58.24 & 66.67 & 76.19 & 77.66 & 86.81 & 89.38 & 89.01 & 88.64 & N/A \\ 
    \cdashline{2-12}
    \rule{0pt}{2.2ex} & \multirow{4}{*}{OntoNotes} & MUC & 12.66 & \07.58  &  13.21  & \08.29 & 10.31 &  31.80   & 33.56 &   56.32 & 77.26 \\
    \rule{0pt}{2.2ex} && B$^3$ & 53.80 &  52.26  &  53.54  &  52.41    & 52.20 &   58.43    & 58.66 &   68.20 & 73.43 \\
    \rule{0pt}{2.2ex} && CEAF$_{\phi4}$ &  31.09  &  29.49  &  31.40  &   30.10   & 32.63 &   38.00   & 39.27   & 50.72 & 74.46 \\
    \rule{0pt}{2.2ex} && Avg. F1 &  32.52    &  29.78  &  32.72 &   30.27   & 31.71  &  42.74     & 43.83 &   58.41 & 76.03 \\ 
    \midrule
    % \hline
    \rule{0pt}{2.2ex}\multirow{1}{*}{DST} & MultiWOZ & JGA & 11.11 & 27.96 & 26.61 & 28.08 & 32.30 &  28.12     & 42.24 &  57.40 & 63.79 \\ 
    % \cline{2-11} 
    % \rule{0pt}{2ex} & SGD & JGA &  &  &  &  &  &  &  &  &  \\ 
    \midrule
    % \hline
    \rule{0pt}{2.2ex}Disc. & PDTB-3 & Acc & 10.04 & 10.04 & 10.04 & 16.15 & 17.16 & 26.01 & 39.77 &   43.83 & 76.23 \\ 
    \midrule
    % \hline
    \rule{0pt}{2.2ex}\multirow{10}{*}{QR} & \multirow{2}{*}{MuDoCo} & BLEU & \00.46 & \00.36 & \07.02 & 49.20 & 41.12 & 61.15 & 66.51 &   57.14 & 80.31 \\
    \rule{0pt}{2.2ex} && ROUGE & \01.52 & 12.18 & 10.98 & 65.61 & 56.07 & 74.78 & 77.88 &   79.37 & 92.01 \\ 
    % \cline{2-2}\cdashline{3-12} 
    \cdashline{2-12}    
    \rule{0pt}{2.2ex} & \multirow{2}{*}{QReCC} & BLEU & \04.53 & 31.27 & 26.35 & 40.09 & 28.19 & 38.64 & 58.68 &  55.24 & 58.67  \\
    \rule{0pt}{2.2ex} && ROUGE & 13.91 & 58.18 & 53.10 & 68.32 & 48.27 & 56.40 & 78.74 & 79.98 & 81.75 \\
    % \cline{2-2}\cdashline{3-12} 
    \cdashline{2-12}
    \rule{0pt}{2.2ex} & \multirow{2}{*}{InCar} & BLEU & \00.00 & \07.66 & 12.71 & 27.42 & 28.20 & 42.13 & 48.58 &   63.66 & 88.45 \\
    \rule{0pt}{2.2ex} && ROUGE & \03.41 & 28.76 & 30.45 & 49.63 & 49.96 & 56.73 & 64.18 &   83.51 & 95.24 \\
    % \cline{2-2}\cdashline{3-12} 
    \cdashline{2-12}
    \rule{0pt}{2.2ex} & \multirow{2}{*}{GECOR} & BLEU & \00.20 & 26.40 & 26.32 & 49.99 & 53.27 & 66.30 & 73.80 &   63.34 & 82.56 \\
    \rule{0pt}{2.2ex} && ROUGE & \04.06 & 42.13 & 42.57 & 65.89 & 69.23 & 80.99 & 86.03 &   79.00 & 92.63 \\ 
    % \cline{2-2}\cdashline{3-12} 
    \cdashline{2-12}
    \rule{0pt}{2.2ex} & \multirow{2}{*}{CANARD} & BLEU & \02.61 & 19.39 & 24.24 & 34.66 & 21.34 & 29.32 & 47.24 &   47.12 & 57.46 \\
    \rule{0pt}{2.2ex} && ROUGE & \09.82 & 45.63 & 49.36 & 62.73 & 38.17 & 46.61 & 69.73 &   74.61 & 81.06 \\
    \bottomrule
    % \hline\hline
    \end{tabular}
    
    \caption{Few-shot results of two open-sourced models and GPT-3.5 on the context understanding benchmark. The results with the best number of few-shot examples are reported for each task. Fine-tuning (FT) results serves as a reference when evaluating LLMs' capability under ICL setup.}
    \label{tab:dense}
\end{table*}

\section{Experiments}

\subsection{Implementation Details}
Evaluation was conducted on a computational infrastructure comprising 8 $\times$ A100 GPUs. 
We experiment with three model families. For smaller models, we consider OPT \cite{zhang2022opt}, ranging from 125M to 2.7B. Although OPT also offers larger models, we opt for LLaMA \cite{touvron2023llama} as the mid-sized LMs, spanning from 7B to 65B parameters, due to showcased superior performance by prior works. For large-scale LMs, we leverage \texttt{GPT-3.5-turbo}\footnote{\url{https://platform.openai.com/docs/models/gpt-3-5}}.
For each model, on every dataset, we assess five different settings: zero-shot, one-shot, 5-shot, 8-shot, and 10-shot. We randomly select the examples from the training set for the few-shot prompting.\footnote{WSC273 itself is a test set and thus has no fine-tuning results. We only report the zero-shot results.}

\subsection{Dense Model} \label{subsec:dense_model_res}
% \andy{andy todo: (1) add FT for dst and qr with explanation of sources; (2) run all missing numbers, including 65B if possible; (3) refine table format; (4) explain sota from different sources; (5) 65B issue and explain}
% \andy{65B issue and explain?}

\begin{figure}[t!]
    \centering
    \includegraphics[width=0.49\textwidth]{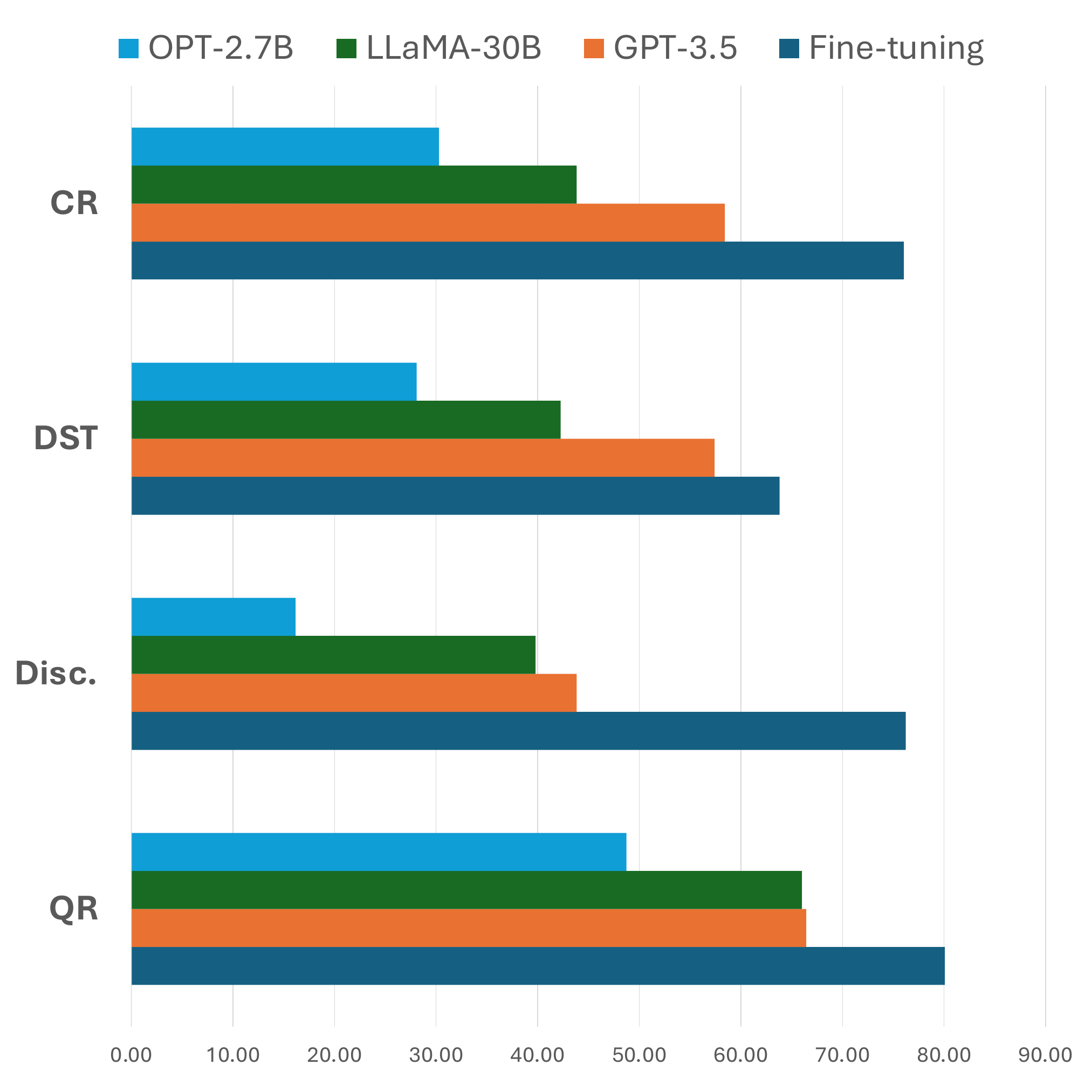}
    \caption{Comparison between commercial/non-commercial models and fine-tuning models for each task in the context understanding benchmark.}
    \label{fig:compare}
\end{figure}

Results of the three model families are reported in Table \ref{tab:dense}, along with results of fine-tuned (FT) models to help better interpret how well the pre-trained models behave with ICL. Figure \ref{fig:compare} also visualizes the gap between various commercial/non-commercial language models and fine-tuning models that achieve the best performance on these tasks. 
% Table \ref{tab:dense} provides the an overview of the benchmark results from three model families and fine-tuning models
% models of varying sizes.
For each, we present the N-shot setting that yields the highest score (see Appendix \ref{appendix:fewshot} for details).
Overall, performance improves as the model size increases and pre-trained models with ICL struggle to catch up with FT models on most tasks.

% Overall, these scores reveal the challenges that LLMs encounter in all four tasks, particularly in the case of smaller models. This trend becomes even more apparent when comparing their performance to the SOTA fine-tuning (FT) results presented in the last column of the table.
% \andy{would be v good to add an column listing all SOTA FT performance (I can help with this), for QR we can use our internal investigation results if no SOTA reported lately, let's check for other three tasks}
% \yz{Done}

\paragraph{Coreference Resolution}Larger models exhibit promising performance on the WSC273 task, indicating that LLMs can effectively handle "simple" coreference relations within limited contexts and mentions. However, when it comes to document-based CR with complex clusters, their performance substantially drops
% \footnote{All models are tested on a subset of OntoNotes with 10\% test data due to the large data size and time-consuming inference.}
\footnote{Note that the OntoNotes dataset is substantially larger than the others. We observe that inference on the entire test set becomes extremely time-consuming, particularly with the larger models; further, the cost of running inference on GPT-3.5 starts becoming non-negligible. Consequently, we propose limiting the OntoNotes test set to a 10\% sub-sample, which is the setting we consistently adopt.}.
Even on providing ground-truth mentions,
% \jm{"labeled" instead of "markable"? or even just "gold spans"?}
% \yz{Done. Use `gold mention span` instead.}
the highest-performing GPT is only on par with rule-based coreference systems \cite{manning-EtAl:2014:P14-5} and is far from the end-to-end fine-tuned SpanBERT \cite{joshi-etal-2020-spanbert}.
% fine-tuned T5 \cite{bohnet-etal-2023-coreference}.
% \andy{Yilun todo: (1) replace FT results and citation}
% \yz{Done}
The gap between ICL and FT results highlights that under the ICL setting, LLMs struggle to build coreference chains without adequate domain-specific examples.
Specifically, models except GPT perform significantly worse on the MUC metric. Error analysis reveals that these models are inclined to create more clusters, including singleton clusters. This implies that pre-trained LLMs encounter difficulties in understanding long-range contextual information.
% \andy{as mentioned, SOTA FT could help establish another upper bound to show the gap between ICL and FT}
% \yz{Done}
% Our error analysis further elucidates
% \yz{TODO: Insert details from error analysis here}
% \andy{And insights from low numbers in OntoNotes?}
% \yz{Done}

\paragraph{DST} A similar trend is observed as CR where OPT and LLaMA models fall behind GPT-3.5 significantly. This suggests that these models fail to extract key information as the conversation proceeds, even with the provision of 5 to 10 demonstrations and the distilled relevant domain ontology in prompt. Our error analysis indicates that most of the errors happen due to the misdetection of slots or the wrong predicted value in a slot-value pair. Only GPT-3.5 reaches the level of FT results which is a fine-tuned T5 base model \cite{bang-etal-2023-task}.

% DST results indicate that smaller models
% % , when compared to ChatGPT in \citet{heck-etal-2023-chatgpt},
% % \andy{what does this comparison mean?}
% % \yz{Done. Remove it}
% are capable of discerning a certain level of dialogue states. However, it is evident that model size limits performance, as shown in OPT -- when increasing the model size to beyond 13B 
% % \andy{5B?}
% % \yz{Wait for 13B results}
% parameters, there is a notable and consistent increase in the JGA score corresponding to the model's size, as observed in the LLaMA results. In addition, to the best of our knowledge, GPT-4 achieves the highest ICL performance among all LLMs, which is also near comparable with the SOTA model (63.79 by \citet{bang-etal-2023-task}). This highlights that LLMs with few-shot examples as prompts can capture information from the current user-system interaction and update the previous state.
% \andy{andy todo: rewrite according to ft dst results}
% \andy{60 by GPT-4 is actually v high, let's see what's the SOTA now and update this statement correspondingly}
% \yz{Rewrite the findings}

% \jm{"Surprisingly,"? Or maybe make the next sentence: "For instance, rather surprisingly, OPT 125M outperforms LLaMA 13B by a significant margin."}
% \yz{Done. Rewrite the whole paragraph}
% \andy{we remove this statement}
% With respect to the implicit discourse relation classification task, the results do not exhibit a consistent alignment with model sizes. For instance, OPT 125M is unexpectedly outperforming LLaMA 13B by a significant margin.

\paragraph{Implicit Discourse Relation Classification}We observe an increase in scores when the model size exceeds 7B. However, even the best-performing LLM, GPT, performs worse than the SOTA fine-tuned model \cite{liu-strube-2023-annotation} with the drop of 32\% accuracy.
% Another observation is that three OPT models (125M, 350M, 1.3B) all achieve the same accuracy score.
% \andy{how? it's 10.04 vs 26.01 shown in table}
% \yz{Done. Rewrite the whole paragraph}
We carefully examine the predictions for each model and found that all models tend to predict the same relation class for every example, albeit with their individual preferences for the selected relation. In addition, because of an imbalanced distribution of classes, these models potentially perform worse than random chance (25\%).
This suggests that the models struggle to distinguish the nuances between different relation classes and fail to correctly identify relations across EDUs within context.
% These results and findings demonstrate the persistent challenge of understanding discourse relations within an ICL setting.

\paragraph{Query Rewriting}The gap between small and large models is significantly huge, compared to the other tasks. For instance, OPT-125M cannot even complete the rewriting task. Analysis on predictions of small models indicates that the model is not capable of following the instructions or learning patterns from the few-shot examples. We identify a few major error types: (1) generating the next sentence, instead of rewriting; (2) rewriting the wrong user turn from the conversation; (3) copying the last user utterance without any rewriting. These errors get reduced as the model size increases. However, similar to the previous three tasks, the best ICL results achieved by GPT is far from the fine-tuned models.\footnote{In literature, the best FT results come from different models across five QR datasets, where some are not even LLM based. To ensure fair comparison, we fine-tuned a T5 large model on each QR dataset.}
It is worth noting that OPT-2.7B performs on par or notably better than LLaMA-7B, which is somewhat not aligned with the findings in \citet{open-llm-leaderboard} where LLaMA-7B even outperforms OPT-66B in many tasks, including ARC \cite{clark2018think}, HellaSwag \cite{zellers2019hellaswag}, and MMLU \cite{hendrycks2021measuring}.

% Among the five query rewriting corpora, LLMs exhibit three consistent trends across each dataset.
% First, within the same model family, performance improves as model size increases.
% an increase in model size generally leads to improved precision and recall scores, which also aligns with the findings from the other three tasks.
% \andy{either cite or remove the sentence}
% \yz{Done.}
% Second, despite \citet{open-llm-leaderboard}'s demonstration that LLaMA-7B outperforms OPT-66B in most evaluation datasets, including ARC \cite{clark2018think}, HellaSwag \cite{zellers2019hellaswag}, and MMLU \cite{hendrycks2021measuring}, OPT-2.7B surpasses LLaMA-7B in terms of BLEU and ROUGE-1 scores in three of five datasets (with two datasets achieving comparable scores).
% \jm{It actually seems to outperform only on 3 of the 5? Though it performs comparably on the fourth?}
% Third, when the model size increases to 30B or beyond, a significant increase on both BLEU and ROUGE is observed, showing that the increase of parameters contribute to the understanding of context.
% However, the best ICL results achieved by GPT is far from fine-tuned models xxx.
% \andy{andy todo: fill the number here}
% \andy{not sure if this holds true, LLaMA-30B and GPT-4 does outperform significantly OPT-2.7B and even smaller models}
% \yz{Done. Rewrite the sentence}

All in all, this section presents a holistic comparison of LLMs' behaviors on the target context understanding tasks. On the tasks with structured outputs such as CR or DST, even small models show a certain level of context understanding and seem to follow the task instruction. Classification tasks such as discourse relation selection are deemed the easiest among all tasks; however, the small models are even worse than a random guess (25\%).
% \andy{confirm with Yilun how to do inference exactly}
As for the generative task, the ability to complete query rewriting can be only observed in the case of larger models, as the model has the freedom to generate arbitrary content that does not follow the prompt.
We notice that OPT-2.7B outperforms LLaMA-7B in multiple QR datasets, including MuDoCo, QReCC, and CANARD. We carefully compare the outputs between the two models. As an example, QReCC, a QA-based conversational dataset, consists of several QA pairs as context and a last query to be rewritten. We observe that LLaMA-7B tends to rewrite the question in context instead of rewriting the last target query, which is not frequent in OPT-2.7B.
It is also noted that except for DST, FT models demonstrate marked superiority over pre-trained models, highlighting the potential for improving LLMs' competence on these context understanding tasks.

% our contextual comprehension benchmark provides a different perspective in the assessment of
% % \jm{"to evaluate" -> "in the assessment of"?}
% LLMs compared with previous evaluation. 
% It shows that the increase of model size (especially beyond 30B) can significantly enhance its contextual comprehension capability.
% It also shows that the models' performances are not consistent with other benchmarks, further demonstrating the necessity of our newly proposed benchmark.
% % \andy{Do we answer or highlight all these points: (1) are SOTA LLMs capable of understanding different properties required by different datasets? (by column) (2) when numbers are low, any insights behind? is it b/c the task itself is difficult? or b/c of our prompt design? or b/c of retrieved examples or anything else? (3) best ICL with biggest model can reach FT level?}
% \andy{andy todo: refine this paragraph to provide a overall summary}

\subsection{Model Compression Technique}
As we focus on evaluating context understanding of LLMs in an ICL setup, we evaluate models quantized using GPTQ \cite{frantar-gptq}, which is an efficient one-shot weight quantization algorithm based on approximate second-order information that compresses the model post-training.
It enables a reduction in memory and disk requirements by up to 80\%, compared to the pre-quantized model.
% \andy{confirm with Yilun 80\% from paper?}

\subsection{Quantized Model Results}
\begin{table}[]
    \centering\small
    \aboverulesep=0ex
    \belowrulesep=0ex
    \begin{tabularx}{0.48\textwidth}{>{\centering\arraybackslash}X|c|c|c|c}
    \toprule
    % \hline
    \rule{0pt}{2.2ex}Dataset & Metrics & 7B-D & 30B-Q & 30B-D \\
    \midrule
    % \hline
    \rule{0pt}{2.2ex}WSC273 & Acc & 86.81 & 87.18 & 89.01 \\ 
    \cdashline{1-5}
    \rule{0pt}{2.2ex}\multirow{4}{*}{OntoNotes}   & MUC            & 10.31 & 25.37 & 33.56 \\ 
    \rule{0pt}{2.2ex}                              & B$^3$          & 52.20 & 56.80 & 58.66 \\ 
    \rule{0pt}{2.2ex}                              & CEAF$_{\phi4}$ & 32.63 & 36.93 & 39.27 \\ 
    \rule{0pt}{2.2ex}                              & Avg. F1        & 31.71 & 39.70 & 43.83 \\ 
    \midrule
    \rule{0pt}{2.2ex}MultiWOZ & JGA            & 32.30 & 41.99 & 42.24 \\ 
    \midrule
    % \rule{0pt}{2.2ex} & SGD                          & JGA            &  &  &  \\ \hline
    \rule{0pt}{2.2ex}PDTB-3              & Acc            & 17.16 & 31.29 & 39.77 \\ 
    \midrule
    % \hline
    \rule{0pt}{2.2ex}\multirow{2}{*}{MuDoCo} & BLEU & 41.12 & 59.22 & 66.51 \\ 
    \rule{0pt}{2.2ex}                              & ROUGE         & 56.07 & 71.38 & 77.88 \\ 
    \cdashline{1-5}
    \rule{0pt}{2.2ex}\multirow{2}{*}{QReCC}       & BLEU           & 28.19 & 53.72 & 58.68 \\ 
    \rule{0pt}{2.2ex}                              & ROUGE         & 48.27 & 74.13 & 78.74 \\ 
    \cdashline{1-5}
    \rule{0pt}{2.2ex}\multirow{2}{*}{InCar}       & BLEU           & 28.20 & 39.69 & 48.58 \\ 
    \rule{0pt}{2.2ex}                              & ROUGE         & 49.96 & 56.32 & 64.18 \\ 
    \cdashline{1-5}
    \rule{0pt}{2.2ex}\multirow{2}{*}{GECOR}       & BLEU           & 53.27 & 70.41 & 83.36 \\  
    \rule{0pt}{2.2ex}                              & ROUGE         & 69.23 & 73.80 & 86.03 \\ 
    \cdashline{1-5}
    \rule{0pt}{2.2ex}\multirow{2}{*}{CANARD}      & BLEU           & 21.34 & 45.07 & 47.24 \\ 
                                  & ROUGE         & 38.17 & 67.15 & 69.73 \\     
    \bottomrule
    % \hline
    \end{tabularx}
    
    \caption{Comparison between dense and quantized models. Dense LLaMA-7B and 3-bit quantized LLaMA-30B share similar memory and disk requirements. \textbf{D} represents dense model and \textbf{Q} denotes quantized model.}
    \label{tab:dense_quantized}
\end{table}

% \andy{Yilun todo: try single column table?}
% \andy{Can move to sec4.4}\yz{Done}
% \andy{talk about quantization setup before any numbers? w/ Joel}
GPTQ \cite{frantar-gptq} has been shown to effectively reduce the model size to 3 bits without incurring substantial performance losses across a range of NLP tasks, such as MMLU, ARC, StoryCloze. However, whether this performance preservation can be extended to contextual understanding was unclear.
% enables us to significantly reduce the model size down to 3 bits without incurring substantial performance losses. This quantization approach allows us to reduce memory and disk requirements by up to 80\% compared to using the dense model.
% \andy{refine the 2 sents - sth like citep shows that quatization allows the model size shrinked significantly on the task of ABC, However, it's unclear whether this holds true for contextualization understanding.}
% \yz{Done}
% For example, in the case of LLaMA, a 3-bit quantized model with 30B parameters exhibits resource requirements that are comparable to a dense model with only 7B parameters.
% \andy{are we sure that llama is fp16? If so, let's make it clear $30*3/16=5.62$}

Table \ref{tab:dense_quantized} presents the comparison between the dense and 3-bit quantized LLaMA models.
% \andy{where we describe the choice of 7B vs 30B?}
In contrast to previous studies on 3-bit quantization, we observed that quantization leads to fluctuated drops in performance across the four tasks. Specifically, in WSC273, MultiWOZ, and CANARD, post-training quantization incurs only a marginal performance drop ($\sim$1.7 points). However, in the remaining datasets, quantization results in significant performance drops.
% Similar to prior research, we also observe that 3-bit quantization only results in a slight drop in performance compared to the dense model.
% \andy{not exact, drop actually fluctuates and could be large across different task; also very important - this statement is not aligned with abstract!}
% \yz{Done}
% \andy{high level talk about Loss in 30Q vs 30D first, how's that compare to the paper? then 30Q vs 7D}
% \yz{Done}

The results further show that the quantized LLaMA-30B model consistently outperforms the dense LLaMA-7B model across all tasks despite being comparable in disk and memory requirements.
For CR, the 30B quantized model achieves significantly higher scores on the OntoNotes dataset across all metrics. The MUC metric shows the most substantial improvement, indicating that the quantized 30B model partially overcomes the tendency to create small clusters for mentions.
% The gains range from 4 to 15 points of improvement over the dense 7B model.
For DST on MultiWOZ, the 30B quantized model show a 30\% relative improvement over the 7B model in JGA.
On discourse parsing with PDTB-3, the accuracy of quantized 30B model is almost double, 17.16\% vs 31.29\%.
Across all QR datasets, the quantized 30B model substantially improves NLG scores compared to the dense 7B model, with relative gains ranging from 15-50\%. The largest gap is observed on GECOR.
% The largest improvements are seen on the GECOR dataset, where the 30B quantized model achieves BLEU and ROUGE scores that are over 20 points higher than the 7B dense model.

In general, we show that the quantized 30B LLaMA model consistently and significantly outperforms the dense 7B model as a result of the increased scale, despite using 3-bit quantization. The benefits of greater model scale thus outweigh the impacts of quantization in understanding discourse. We believe this finding would be beneficial when deploying LLMs in real-world applications with disk and runtime constraints.

\section{Case Study: Query Rewriting}
% As mentioned in Sec \ref{subsec:dense_model_res} we found the small models are not able to perform query rewriting successfully by several reasons. Fundamentally it is a sentence generation task and the output space is much larger comparing to the other three tasks with structured outputs. We therefore take query rewriting as a case study, hoping to gain more insights behind LLMs' generation with a more thorough analysis.
% So far we have shown results of a variety of models spanning model sizes and the impact of quantization on LLaMA.
In this section, we provide in-depth analysis by comparing the two open-sourced model families OPT and LLaMA, and the impact of quantization, using query rewriting as the target task.

We conduct a careful inspection of the query rewriting task because of three reasons: (1) by the nature of the task, query rewriting is the only one with free-form generation, while the others effectively are either classification-based tasks or heavily constrained in their possible output predictions. The generation task allows us to explore the LLMs' output in more detail, and to provide more interesting insights; (2) the manual analysis of errors is a time-consuming process, making it challenging to conduct such an in-depth analysis across all four tasks; (3) the query rewriting task covers a diverse range of five datasets, enabling us to compare differences between each dataset and to thereby gain a deeper understanding.

% In this section we aim to provide more insights when comparing the two (open sourced) model families and the impact of quantization. We take query rewriting as the target task for analysis.
% on the query rewriting task.
% In this section we aim to provide more insights 

% In this section we provide a more detailed analysis on this task and hope to provide more insights.
% \andy{I rewrite here and the beginning of each subsection}

% In this section, we use the query rewriting task as a case study to comprehensively examine the contextual understanding capabilities of the models under investigation
% \jm{I'm not sure we need the reason why, it kinda draws (possibly negative) attention to the fact that the other tasks don't have as much data maybe?}.
% \yz{Change to a neutral expression}
% First, we compare the performance of OPT-30B and LLaMa-30B, showing a surprising contrast between our observation of how the models perform and that of prior work. Next, we conduct a detailed qualitative analysis comparing dense and quantized models, and the error patterns observed in each, in the context of query rewriting. \jm{Yilun, Andy, could you please review? Took a crack at writing this, but possibly needs some polishing.}

% \andy{probably don't need Sec5: Analysis section - both subsections can be moved to Sec4; Sec5.2 renamed to Case Study: query rewriting}
% \andy{Sec5 title -> Case Study: query rewriting}
% \andy{if space allowed, add why we select QR for case study - discuss with Joel}

\subsection{OPT vs. LLaMA}
\begin{table}[t!b]
    \centering\small
    \aboverulesep=0ex
    \belowrulesep=0ex
    \begin{tabular}{l|cc|cc|cc}
    \toprule
    % \hline
    \rule{0pt}{2.2ex}\multirow{2}{*}{Dataset}                     & \multicolumn{2}{c|}{6.7/7B}       & \multicolumn{2}{c|}{13B}          & \multicolumn{2}{c}{30B} \\
    \rule{0pt}{2.2ex} & O. & L. & O. & L. & O. & L. \\ 
    \midrule
    \rule{0pt}{2.2ex}\multirow{2}{*}{Mudoco} & 53.1 & 41.1 & 55.2 & 61.1 & 55.2       & 66.5       \\
    \rule{0pt}{2.2ex}                        & 71.8 & 56.0 & 72.1 & 74.7 & 71.5       & 77.8       \\ 
    \hdashline
    \rule{0pt}{2.2ex}\multirow{2}{*}{QReCC}  & 46.6 & 28.1 & 43.7 & 38.6 & 43.8       & 58.6       \\
    \rule{0pt}{2.2ex}                        & 73.4 & 48.2 & 71.6 & 56.4 & 71.9       & 78.7       \\ 
    \hdashline
    \rule{0pt}{2.2ex}\multirow{2}{*}{InCar}  & 40.3 & 28.2 & 41.9 & 42.1 & 44.6       & 48.5       \\
    \rule{0pt}{2.2ex}                        & 64.8 & 49.9 & 62.6 & 56.7 & 65.3       & 64.1       \\ 
    \hdashline
    \rule{0pt}{2.2ex}\multirow{2}{*}{GECOR}  & 58.8 & 53.2 & 60.9 & 66.3 & 58.2       & 73.8       \\
    \rule{0pt}{2.2ex}                        & 75.7 & 69.2 & 78.3 & 80.9 & 76.1       & 86.0       \\ 
    \hdashline
    \rule{0pt}{2.2ex}\multirow{2}{*}{CANARD} & 43.8 & 21.3 & 37.5 & 29.3 & 41.3       & 47.2       \\
    \rule{0pt}{2.2ex}                        & 72.0 & 38.1 & 66.0 & 46.6 & 69.3       & 69.7      \\
    \bottomrule
    % \hline
    \end{tabular}
    \caption{Comparison between OPT (O.) and LLaMA (L.) across five query rewrite datasets. For each dataset, the first and second rows represent BLEU and ROUGE scores respectively.}
    \label{tab:opt_llama_30b}
\end{table}
% \andy{andy todo: peep into predictions and refine this subsec}
% \andy{andy todo: (1) avg 2 runs; (2) run OPT-7b, 13b; (3) send llama-7b canard}
% \andy{this can be cut if running out of space as it requires an extra table (appendix?)}
% We first compare OPT and LLaMA under the same model size.
Prior works \cite{open-llm-leaderboard} have consistently shown that, under the same model size, LLaMA outperforms OPT. However, their performance on QR, as shown in Table \ref{tab:opt_llama_30b}, does not follow this pattern.

When the model size is around 7B, OPT consistently performs better than LLaMA by a significant margin across the five QR datasets. The two models perform on par with each other at 13B. The superiority of LLaMA is only obvious with 30B model size.
From another perspective, although we expect performance to improve as model size increases, we observe this trend on LLaMA, but not on OPT.
These results suggest that it may not be correct to conclude the overall superiority between two model families by only comparing on a certain range of model sizes or on a certain set of tasks.

\begin{table}[t!b]
    \centering\small
    \aboverulesep=0ex
    \belowrulesep=0ex
    \begin{tabular}{p{0.45\textwidth}}
        \toprule
        % \hline
        \rule{0pt}{2.2ex}\textbf{Context} \\
        \texttt{User}: what is the name of india pakistan border line \\
        \texttt{Bot}: The Radcliffe Line was the boundary demarcation \\ 
        \hspace{1mm} line between the Indian and Pakistani portions of the \\ \hspace{1mm} Punjab and Bengal provinces of British India. \\
        \texttt{User}: who created the radcliffe line \\
        \texttt{Bot}: The Radcliffe Line was named after its architect, Sir \\ \hspace{1mm} Cyril Radcliffe, who was the joint chairman of the two \\ \hspace{1mm} boundary commissions for the two provinces. \\
        \texttt{User}: when was the line published \\
        \midrule
        % \hline
        \rule{0pt}{2.5ex}\textbf{Gold answer}: when was the \underline{radcliffe} line published \\
        \midrule
        % \hline
        \rule{0pt}{2.5ex}\textbf{Prediction 1 (repeat the last query)}: when was the line published \\
        \midrule
        % \hline 
        \rule{0pt}{2.5ex}\textbf{Prediction 2 (language modeling)}: 1947 \\
        \bottomrule
        % \hline
    \end{tabular}
    \caption{An example of two major types of errors found in the query rewriting task. 
% \andy{unclear by using the name of the error types: it's hard to follow and understand "1947" is the model prediction. Suggest -> Prediction (repeating etc): ...}
}
    \label{tab:qr_err}
\end{table}

\subsection{Dense vs. Quantized}
We conduct a quantitative analysis on the error types of query rewriting to investigate the performance gap between dense and quantized models. Across the five datasets, we identify two main error types that account for nearly 80\% of the total errors, with examples shown in Table \ref{tab:qr_err}.
% \andy{positions - tb all tables/figures}
% \yz{Done}
% \andy{switch the two error types (probably with better names?) as technically the first error does not follow the instruction as well}
% \yz{Done}
First, the model \textit{repeats} the last query without resolving any referred entity or ellipsis. In this case, the model seems to understand the instruction but fails at rewriting. This type of error can be primarily associated with the model's context understanding capability.
Second, the model treats the task as a language modeling (\textit{LM}) task, where it provides a response to the last query. In this scenario, the model appears to struggle to understand the task instruction, even with several few-shot examples. We classify this error type as more related to the model's ICL ability.

\begin{table}[t!b]
    \centering\small
    \aboverulesep=0ex
    \belowrulesep=0ex
    \begin{tabularx}{0.48\textwidth}{X|P{13mm}|Y|Y|Y}
        \toprule
        % \hline
        \rule{0pt}{2.2ex}Type & Dataset & 7B D & 30B Q & 30B D \\
        \midrule
        % \hline
        \rule{0pt}{2.2ex}\multirow{6}{*}{Repeat} & MuDoCo & 260 & 247 & 194 \\
        \rule{0pt}{2.2ex} & QReCC & \086 & \090 & \026 \\
        \rule{0pt}{2.2ex} & InCar & \017 & \015 & \0\08 \\
        \rule{0pt}{2.2ex} & GECOR & \059 & \062 & \037 \\
        \rule{0pt}{2.2ex} & CANARD & \047 & \044 & \032 \\
        \cdashline{2-5}
        \rule{0pt}{2.2ex} & Total & 469  & 458  & 297  \\
        \midrule
        % \hline
        \rule{0pt}{2.2ex}\multirow{6}{*}{LM} & MuDoCo & \071 & \029 & \016 \\
        \rule{0pt}{2.2ex} & QReCC & \080 & \028 & \016 \\
        \rule{0pt}{2.2ex} & InCar & \019 & \020 & \015 \\
        \rule{0pt}{2.2ex} & GECOR & \0\06 & \0\01 & \0\00 \\
        \rule{0pt}{2.2ex} & CANARD & 127 & \076 & \059 \\
        \cdashline{2-5}
        \rule{0pt}{2.2ex} & Total & 232 & 125 & 106 \\
        \bottomrule
        % \hline
    \end{tabularx}
    \caption{Number of the major two types errors on three LLaMA models (7B dense, 30B quantized, and 30B dense) found in query rewriting. \textit{Repeat} stands for repeat-the-last-query error and \textit{LM} denotes language modeling error.
    % \andy{consistency in error type names}
    % \yz{Done}
    % \andy{remove CANARD and add total column?}
    % \yz{Done}
    % \andy{add total numbers?}
    % \yz{TODO: correct numbers}
    % \andy{Yilun to double check Total}
    }
    \label{tab:qr_numbers}
\end{table}

We perform manual error annotations on the five QR datasets\footnote{10\% test data on QReCC and CANARD was graded.}.
% including MuDoCo, QReCC\footnote{For QReCC, we select 10\% of the test data.}, InCar, and GECOR.
% We perform manual error annotations on the five datasets, with the exception that for QReCC and CARNARD 
% \andy{can just say we did on A, B and C \jm{+1}}
% \yz{Done}
% \andy{only CANARD is left out in table?}
% \yz{Done}
% , we selected 10\% of the test data for error analysis. 
% \andy{make it a footnote and clarify it's only for qrecc}
% \yz{Done}
Table \ref{tab:qr_numbers} illustrates the number of errors of the three selected models on each dataset.
% \andy{results are just not filled in?}
A consistent trend is observed across all QR datasets. 
In terms of \textit{repeat} errors, the 30B dense model exhibits significantly fewer errors compared to the 7B dense model (297 vs. 469). However, 3-bit GPTQ quantization leads to an increase in this type of error, reaching a similar error count to the 7B dense model (458 vs. 469). This implies that 3-bit quantization reduces the model's ability to comprehend the context.
% \jm{Can we just say "ability to comprehend context"?}
% \yz{Done}
Regarding \textit{LM} errors, the 30B dense model also significantly outperforms the 7B dense model, with 106 errors compared to 232. It is to be noted that the quantized model generates only 125 \textit{LM} errors,
% \jm{Summing up the 30B Q QA column gives me 110 errors, not sure what I'm missing?}
% \andy{+1, a row with sum would help}
% \yz{TODO: correct numbers}
slightly more than the 30B dense model. However, it generates significantly fewer (around 50\%) errors compared to the 7B dense model (125 vs. 232).
% \jm{Consider rewording "representing less than 15\% of the error count" to "making only 15\% of the errors of the 7B dense model" (if I understood this correctly?)}
% \yz{Done}
This indicates that 3-bit quantization maintains the ICL capability that allows models to rewrite the user query successfully rather than performing language modeling task.

\section{Conclusion}
This paper introduces a contextual understanding benchmark designed to assess the performance of LLMs. We collect nine 
% \andy{change all 10 to 9 across paper} 
% \yz{Done}
existing datasets spanning four tasks, each carefully tailored to suit generative models. This benchmark encompasses essential elements for assessing linguistic comprehension within context, including both document and dialog based contextual understanding. Experimental results reveal that LLMs under in-context learning struggle with nuanced linguistic features within this challenging benchmark, exhibiting inconsistencies with other benchmarks that emphasize other
% \jm{"other"?} 
% \yz{Done}
aspects of language. To the best of our knowledge, we are also
% \jm{"also"}
% \yz{Done}
the first to compare dense models and post-training quantization models in contextual understanding tasks.
% \andy{not precise - we are the first comparing these two in context understanding tasks}
This comparison highlights that 3-bit post-training quantization reduces the general understanding capacity of context to different extent across the 4 tasks. The proposed contextual comprehension benchmark thus provides a unique perspective on the contextual dimension of language understanding and offers a valuable addition to existing LLM evaluations.

\section*{Limitations}
This work provides an evaluation of various pre-trained LLMs, including OPT, LLaMA, and GPT, on our understanding benchmark. However, we have not evaluated other LLMs designed for longer input scenarios, such as LongLLaMA \cite{tworkowski2023focused}.

In our evaluation, we focus on the GPTQ quantization method, analyzing its performance on our benchmark. We do not include other post-training quantization techniques, such as RPTQ \cite{yuan2023rptq}, in this work.

Our evaluation concentrates on English datasets, primarily utilizing LLMs pre-trained with English data. All of the four tasks on our benchmark have datasets from other languages. The coreference dataset OntoNotes 5.0 contains annotations of Arabic and Chinese. In addition, recent releases such as CorefUD \cite{nedoluzhko-etal-2022-corefud} promote standardization of multilingual coreference annotations. In DST, CrossWOZ \cite{zhu-etal-2020-crosswoz} is a cross-domain wizard-of-oz task-oriented dataset. \citet{long-etal-2020-ted} develop TED-CDB, a Chinese discourse relation dataset. The query rewriting task also has datasets in other languages, such as REWRITE \cite{su-etal-2019-improving} and Restoration-200K \cite{pan-etal-2019-improving}. Finally, specific LLMs optimized for individual languages, such as ChatGLM \cite{du2022glm}, exist and are not a part of our evaluation.

\section*{Acknowledgements}
The authors would like to thank Jeffrey Nichols, Russ Webb and the anonymous reviewers for their help and feedback.

\bibliography{anthology,custom}
% \bibliographystyle{acl_natbib}

% \clearpage
% \newpage

\appendix
\section{Task Design Examples} \label{appendix:task_design}
Table \ref{tab:task_design} presents the input example for each task. For CR, we only show examples from OntoNotes.
\begin{table*}[]
    \centering\tiny
    \aboverulesep=0ex
    \belowrulesep=0ex
    
    \begin{tabular}{p{\textwidth}}
    \textbf{Coreference Resolution}
    \newline 
    \rule{0pt}{2.5ex}Instructions: Please carefully read the following passages. For each passage and the options, you must identify which option the mention marked in *bold* refers to. If the marked mention does not have any antecedent, please select ``no antecedent''.
    \newline\newline
    [Few-shot examples]
    \newline\newline
    Context: — basically , it was unanimously agreed upon by the various relevant parties . To express *its* determination , the Chinese securities regulatory department compares this stock reform to a die that has been cast . It takes time to prove whether the stock reform can really meet expectations , and whether any deviations that arise during the stock reform can be promptly corrected . Dear viewers , the China News program will end here . This is Xu Li . Thank you everyone for watching . Coming up is the Focus Today program hosted by Wang Shilin . Good-bye , dear viewers .
    \newline
    Choice: \newline
    A. the Chinese securities regulatory department \newline
    B. this stock reform \newline
    C. the stock reform \newline
    D. you \newline
    E. everyone \newline
    F. no antecedent \newline
    Question: What does *its* refers to? \newline
    Answer: \textit{A}
    \\ \hline

    \rule{0pt}{2.5ex}\textbf{Dialogue State Tracking}
    \newline
    \rule{0pt}{2.5ex}Consider the following list of concepts, called "slots" provided to you as a json list.\newline

    ``slots'': \{``restaurant-pricerange'': ``price budget for the restaurant'', \newline
    \hangindent=1em``restaurant-area'': ``area or place of the restaurant'',\newline
    ``restaurant-food'': ``the cuisine of the restaurant you are looking for'',\newline
    … \newline
    ``hotel-postcode'': ``postal code of the hotel'',\newline
    `hotel-ref'': ``reference number of the hotel booking'' 
    
    \}\newline
    
    Some ``slots'' can only take a value from predefined list:\newline
    
    ``categorical'': \{``restaurant-pricerange'': [`cheap', `expensive', `moderate'], \newline
    \hangindent=1em ``restaurant-area'': ['centre', 'east', 'north', 'south', 'west'],\newline
    ``restaurant-bookday'': ['monday', 'tuesday', 'wednesday', 'thursday', 'friday', 'saturday', 'sunday'],\newline
    …\newline
    ``hotel-internet'': ['free', 'no', 'yes'],
    ``hotel-area'': [`centre', `east', `north', `south', `west']
    
    \}\newline
    
    Now consider the following dialogue between two parties called the ``system'' and ``user''. Can you tell me which of the ``slot'' was updated by the ``user'' in its latest response to the ``system''? Present the updates in JSON format. If no ``slots'' were updates, return an empty JSON list. If you encounter ``slot'' that was requested by the ``user'' then fill them with ``?''. If a user does not seem to care about a discussed ``slot'' fill it with ``dontcare''.\newline
    
    Input: \newline
    Previous state: \{\}\newline
    ``system'': ``''\newline
    ``user'': ``I'm looking for a moderately priced place to eat that's in the centre of town.''\newline
    Output: \textit{\{``restaurant-pricerange'': ``moderate'', ``restaurant-area'': ``centre''\}}
    \\ \hline

    \rule{0pt}{2.5ex}\textbf{Implicit Discourse Relation Classification}
    \newline
    \rule{0pt}{2.5ex}Instructions: Given two arguments and a list of connective words, please select the most likely connective between two arguments.
    \newline\newline
    Below are the descriptions of four discourse relation labels. Please find the correct label for each example.
    \newline\newline
    Temporal: The tag temporal is used when the situations described in the arguments are intended to be related temporally. \newline
    Contingency: The tag Contingency is used when the situation described by one argument provides the reason, explanation or justification for the situation described by the other. \newline
    Comparison: The tag Comparison is used when the discourse relation between two arguments highlights their differ- ences or similarities, including differences between expected consequences and actual ones. \newline
    Expansion: The label Expansion is used for relations that expand the discourse and move its narrative or exposition forward.
    \newline\newline
    [Few-shot examples]
    \newline\newline
    Input: \newline
    Arg 1: Amcore, also a bank holding company, has assets of \$1.06 billion. \newline
    Arg 2: Central's assets are \$240 million. \newline
    Question: What is the connective that best describes the relation between two arguments? \newline
    A. Temporal \newline
    B. Contingency \newline
    C. Comparison \newline
    D. Expansion \newline
    Answer: \textit{C}
    \\ \hline

    \rule{0pt}{2.5ex}\textbf{Query Rewrite}
    \newline
    \rule{0pt}{2.5ex}Instructions: Rewrite the last query following interaction into a well-formed, context independent query. Resolve any disfluencies or grammatical errors in the query.
    \newline\newline
    [Few-shot examples]
    \newline\newline
    Input: \newline
    User: Try to reach Forbes now . \newline    
    Bot: Forbes at Washington Post ? Or Forbes of Publishing Division ? \newline
    User: Publishing Division . \newline
    Rewrite: \textit{Forbes of Publishing Division}
    \\ \hline
    \end{tabular}
    
    \caption{Examples of task design for each task in the context understanding benchmark.}
    \label{tab:task_design}
\end{table*}

% \section{Experiment Setup}

\section{Few-shot Settings} \label{appendix:fewshot}
Table \ref{tab:few_shot} shows the number of examples for each dataset that yields the best scores. All datasets except WSC273 and PDTB3 use randomly selected examples from the training set. Since WSC273 does not include any train or validation set, we use the zero-shot setting, as scores presented in Table \ref{tab:dense}. For each class in PDTB3, we randomly select two examples from the training set for prompting. For some particular datasets, such as OntoNotes, experiments are only performed in the zero-shot and one-shot settings due to the limitation on input length.

\begin{table*}[]
    \centering\small
    \aboverulesep=0ex
    \belowrulesep=0ex
    \begin{tabular}{l|cc|c|c|ccccc}
    \toprule
    \rule{0pt}{2.2ex}Task & \multicolumn{2}{c|}{Coreference} & DST & Discourse & \multicolumn{5}{c}{query rewriting} \\
    \midrule
    \rule{0pt}{2.2ex}Dataset & WSC273 & OntoNotes & MultiWOZ & PDTB3 & MuDoCo & QReCC & InCar & GECOR & CANARD \\
    \midrule
    \rule{0pt}{2.2ex}N-example & Zero-shot & One-shot & 5-shot & 8-shot & 10-shot & 5-shot & 10-shot & 10-shot & 5-shot \\
    \bottomrule
    \end{tabular}
    \caption{N-shot settings for each task \& dataset that yields the highest scores. For each task and model, we use consistent N-shot settings for comparison.}
    \label{tab:few_shot}
\end{table*}

\section{Reliability of Experiment Results}
For each task, we have randomly run several experimental setups with multiple rounds, with over 10 settings in total. However, due to the challenges posed by limited time, budget, and computing resources, it is very difficult to run multiple rounds for every single experiment, given the complexity of our experimental setup. In addition, for existing experiments with multiple rounds, we empirically observe that there is low variance across the rounds, which leads us to assume that performing the remaining experiments with a single run does not significantly impact the arguments presented in this paper.

% \section{Benchmark Statistics}
% Table \ref{tab:stats} provides statistics for each dataset in our context understanding benchmark.

% \begin{table}[t!bh]
%     \centering\small
%     \begin{tabularx}{0.47\textwidth}{c|c|c|>{\centering\arraybackslash}X}
%     % \toprule
%     \hline
%     Type & Task & Dataset & Number of test \\ 
%     % \midrule
%     \hline
%     \rule{0pt}{0ex}\multirow{3}{*}{Doc} & \multirow{2}{*}{Coreference} & WSC273 & 273 examples \\ \cline{3-4}
%     \rule{0pt}{0ex} && OntoNotes & 348 docs \\ \cline{2-4} 
%     \rule{0pt}{0ex} & Discourse & PDTB-3 & 1474 relations \\ \hline
%     \rule{0pt}{0ex}\multirow{6}{*}{Dial.} & DST & MultiWoz  &  \\ 
%     % \cline{3-3}
%     % \rule{0pt}{2ex} && SGD &  \\ 
%     \cline{2-4} 
%     \rule{0pt}{0ex} & \multirow{5}{*}{\begin{tabular}[c]{@{}c@{}}Query\\ Rewrite\end{tabular}} & MuDoCo & \multirow{5}{*}{Ellipsis and reference}          \\ \cline{3-4}
%     \rule{0pt}{0ex} && QReCC &  \\ \cline{3-4}
%     \rule{0pt}{0ex} && InCar & 214 examples \\ \cline{3-4}
%     \rule{0pt}{0ex} && GECOR &  \\ \cline{3-4}
%     \rule{0pt}{0ex} && CANARD &  \\
%     % \bottomrule
%     \hline
%     \end{tabularx}
%     \caption{Number of test examples for each dataset in the context understanding benchmark.}
%     \label{tab:stats}
% \end{table}

% \section{LLaMA-2 Results}
% \input{tables/llama2}

\end{document}